# A pedestrian path-planning model in accordance with obstacle's danger with reinforcement learning

Thanh-Trung Trinh
Shibaura Institute of Technology
3-7-5 Toyosu, Koto-ku, Tokyo
135-8548 Japan
(+81)50-5339-3233
nb18503@shibaura-it.ac.jp

Dinh-Minh Vu
Shibaura Institute of Technology
3-7-5 Toyosu, Koto-ku, Tokyo
135-8548 Japan
(+81)90-6329-5811
nb17502@shibaura-it.ac.jp

Masaomi Kimura
Shibaura Institute of Technology
3-7-5 Toyosu, Koto-ku, Tokyo
135-8548 Japan
(+81)3-5859-8507
masaomi@shibaura-it.ac.jp

## ABSTRACT
Most microscopic pedestrian navigation models use the concept of "forces" applied to the pedestrian agents to replicate the navigation environment. While the approach could provide believable results in regular situations, it does not always resemble natural pedestrian navigation behaviour in many typical settings. In our research, we proposed a novel approach using reinforcement learning for simulation of pedestrian agent path planning and collision avoidance problem. The primary focus of this approach is using human perception of the environment and danger awareness of interferences. The implementation of our model has shown that the path planned by the agent shares many similarities with a human pedestrian in several aspects such as following common walking conventions and human behaviours.

## 1. INTRODUCTION
Recent studies in pedestrian simulation are often fixated within one of the three categories assorted by the level of interaction: *macroscopic, mesoscopic* and *microscopic* [1]. The macroscopic simulation models use the concept of fluid and particles originated from physics to construct pedestrian navigations while ignoring the interactions between pedestrians as well as individual characteristics of each pedestrian. For an excessively high-density crowd, a macroscopic model could be sufficient; however, for a smaller size of pedestrians where social interactions are essential, a mesoscopic or microscopic model would be more suitable. A mesoscopic model sits between macroscopic and microscopic, which is still able to simulate a relative large-sized environment but with the cost of agent's movements and interactions. Compared to mesoscopic, a microscopic model is more realistic as each pedestrian is considered as an independent object or a computer agent whose behaviours and thinking processes could be modelled upon.

Most microscopic pedestrian simulation models use the concept of "forces" applied to the pedestrian agent to replicate the navigation behaviour [2]. The basic idea of these models is that pedestrian agents are attracted to a specific point-of-interest (e.g. pedestrian's destination) and repulsed from possible collisions (e.g. walls, obstacles and other agents). The representation of the force-based models is similar to the interactions between magnetic objects with some certain improvements. There is undoubtedly a sufficient resemblance in basic movement and collision avoidance with the implementation of the model in a simulation. However, when comparing with pedestrian navigation in real life, many human decisions which require strategical thinking or social interacting are not reflected in the force-based simulation. For instance, when an agent plans a path to go from its current position to a destination, a force-based agent often chooses the shortest path without colliding into other obstacles most of the time. In real life, a human pedestrian has many other aspects affecting his decision such as social comfort, law obeyance or his personal feeling. This could be a problem if the simulation needs the preciseness of pedestrian behaviour, for instance, a traffic simulation system for automated vehicles.

The main idea of our research is adopting reinforcement learning in the pedestrian agent's decision-making process. Reinforcement learning is a machine learning paradigm based on the concept of *reinforcement* in behavioural psychology, in which the learner needs to find an action in the current state for an optimum reward. The concept is virtually close to the way humans learn to behave in many real-life situations, including path navigation. When a person plans a path to the destination and feels uncomfortable with his decision, for instance, because of taking a longer path or colliding with obstacles, he will then receive a negative reward and will try to improve his behaviour. As a result, once an environment is observed, that individual will be able to come up with a path using his current optimum policy without the needs of various calculation such as "forces" realised in many microscopic pedestrian models.

The remainder of this paper is structured as follows: The next section provides an overview of studies related to our research. Section 3 presents the backgrounds of reinforcement learning and the PPO algorithm. After that, we describe the methodology in our path planning model in Section 4. The modelling of our model and the formulation of our rewarding behaviour will be presented in Section 5 and Section 6, respectively. In Section 7, we present the implementation of our model and evaluation with a conventional rule-based model.

## 2. RELATED WORK

One of the most influential algorithms in microscopic pedestrian simulation is the *Social Force Model* (SFM) by Helbing et al. [3]. The concept of this model is that each pedestrian agent will be under influences of different social forces, including driving force, agent interact force and wall interact force. The driving force attracts agent toward the destination, the agent interact force repulses agent from other agents, and the wall interact force repulses agent from walls or boundaries. Since SFM was introduced, there have been a variety of models formulated based on SFM However, such models do not take account of the cognitive thinking process within the human brain, which leads to many deviations from actual human behaviour.

Regarding research in human behaviour, many studies can be found in the field of robotics research. Many researchers have tried to solve the problems in *human comfort* and constructing naturalness [4]. For an agent to navigate naturally, not conflicting with other pedestrians or obstacles is not enough; but the agent also needs to replicate different behaviours from humans. Another concept proposed in human behaviour research is *human bias* or *cognitive bias*, which causes the anomaly in the human decision process. For example, in [5], Golledge et al. have shown that pedestrians do not always choose the most optimised decision while selecting a path. Another study by Cohen et al. [6] also discussed how the human brain making decisions between exploitation and exploration. These aspects were supportive for forming the agent behaviour in our research.

In using reinforcement learning for pedestrian navigation, the amount of research is moderately limited. In a study by Martinez et al. [7], an experiment in using reinforcement learning for a multi-agent navigation system has been implemented; however, the algorithm used was q-learning which is too simple and does not suit well to a dynamic environment. Another approach is learning from observing examples from human behaviour. In their paper by Kretzschmar et al. [8], a navigation model was proposed using inverse reinforcement learning. One difficulty in such approach is the example or the dataset from human behaviour is not easy to be extracted or readily available.

## 3. BACKGROUND
### 3.1 Reinforcement Learning

Reinforcement learning was first introduced by Surton et al. (1998) [9]. A reinforcement learning agent learns to optimise the *policy*, the mapping from a (possibly partial) observed *states* of the environment to *action* to be taken, in order to maximise the expected cumulative *reward*. Different to supervised learning, instead of using existing inputs and outputs, the reward will be given by using the *reward signal*. This could be inferred as a positive or negative experience from humans (such as satisfied or discomfort) in a biological system. However, a positive or negative reward is intermediate, which means that an action is considered bad at that moment but could also yield a better result in the long run. As a result, a reinforcement learning system also needs a *value function* to define the expected long-term reward positivity.

### 3.2 PPO Algorithm

Many modern reinforcement learning algorithms employ different deep learning techniques to optimise the total cumulative reward. These approaches use the neural network training process to optimise the agent's policy. They often calculate an *advantage value* $\widehat{A_t}$ by comparing the expected reward over the average reward for that state. The advantage function will be then used in the loss function of the neural network, which is consequently trained for a number of steps and outputs the most optimised policy.

For algorithm like Policy Gradient, the policy $\pi_\theta(a_t \mid s_t)$ will be constantly updated after every training step. With a noisy environment, the old policy $\pi_{\theta\,old}(a_t \mid s_t)$, which might actually be better than the new one, will be overwritten; causing the training process to be less efficient. The algorithm *Proximal Policy Optimization* (PPO) was introduced by Schulman et al. [10] try to avoid that problem. The objective of the algorithm is to avoid staying away too far from a good policy by keeping the old good policy and compared with updated one using a more efficient loss function.

## 4. PATH PLANNING MODEL

To design our model of pedestrian path planning and collision avoidance using reinforcement learning, we had to address the following problems: the definition of the environment and the formulation of the rewarding behaviour.

Regarding the definition of the environment, one difficulty is that the model of the designed environment cannot be too complicated. If the environment is too complex (e.g. too large or too many obstacles); the agent might not be able to learn the appropriate behaviour, or it could take an excessive amount of time. In addition to that, the environment also needs to provide a stable training process, or the variation between each training states should be balanced. For the former problem, we limited our environment to a relatively small area with an appearance chance of one obstacle, assumed that a more complex walking environment could be divided into smaller paths. For the latter problem, we introduced a mechanism for resetting the environment to balance the proportion between the cases when there is an obstacle present and ones when there is none. For instance, if the agent fails to navigate without collision with an obstacle when there is one; but in the next training step, there is no obstacle so the agent could produce a path without collision. This could make the agent incorrectly thinks that the current policy is a good one, while it is probably not. As a result, instead of resetting every training step, we suggested resetting the environment only when the agent has already planned a path without conflicting with the obstacle. If the agent fails, we retained the states of the environment for a number of iteration before resetting so that the agent could gain enough experience without being stuck in a bad policy.

We also introduced a definition to the obstacle in our environment so that the agents could output a natural path around it. Different to a physical obstacle in real-world (e.g. a rock, a wall or a construction site), the term obstacle in our research represents the feeling of interference while planning a path. For example, a group of people engaging in a conversation in the middle of the walking area could also be considered as an obstacle. Although there could be a considerable amount of possible walking space within the group's territory, planning a path through this area is considered rude or unusual for a normal person. In a study by Chung et al. [11], such areas are called the "spatial effect" in an environment. As the process of constructing a spatial effect area is carried out in the human cognitive system, the interpretation of an obstacle could be slightly different for each person or in different situations (e.g. the crossroad when the traffic light is green or red).

Apart from position, we proposed two properties to our obstacle: *size* and *danger level,* which should have a great impact on the path planned by the agent as suggested in several studies [12]. The size of the obstacle should cover the concept of spatial effect mentioned above, not just the size of the physical obstacle. For example, a damaged or unstable power pole would have a much larger "size" compared to a steady or stable one due to the fear of the pole falling. For simplicity, we assume our obstacle has a round shape; thus, the size of an obstacle will be expressed by a *radius* value. In terms of the danger level, similar to obstacle's size, is also a concept formed within the human mind. The feeling of danger level could have a great impact on the process of planning a path by a pedestrian. For example, in the two settings illustrated in **Figure 1** below, on the left is an obstacle such as a water puddle which has a much less danger level compared to a deep hole on the right. As a result, the planned path would normally stay much further away from the hole than from the puddle. A more detailed description of the obstacle properties was shown in the technical report [13]. The concrete modelling of the environment will be presented in Section 5.

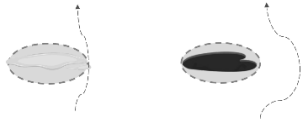

**Figure 1. Path planned by an agent in different settings.**

In addition to the modelling of the environment, we need to specify the appropriate reward function to the agent. For each taken action, the agent needs to know if the action is possibly good or bad based on the given reward. Different from rule-based methods, in reinforcement learning, rewards are often given based on the results of the agent's actions to help the agent in shaping the behaviour. An improper rewarding, for example, giving a large penalty for an undesirable action might cause the agent to avoid such action completely, although there might be a chance that the action could lead to a higher cumulative result in the long run. Another problem with rewarding is that the agent does not receive each specific reward for each behaviour but only the total reward for every action. While this is corresponding to the concept of reinforcement in human cognition, shaping a specific behaviour is much harder compared to in rule-based methods.

As a result, we chose to formalise our rewarding behaviour based on various factors affecting *human comfort* which were summarised in an article by T. Kruse et al. [4]. These are a number of factors applied to robot movement which may cause humans to observe its movement as more natural or human-like. Consequently, a human being should feel the same level of comfort when exhibiting similar behaviour. We will thoroughly present our rewarding formulation in Section 6 of this paper.

## 5. ENVIRONMENT MODELLING

The modelling of our environment is presented as illustrated in **Figure 2**. In the scope of our research, the size of our environment is limited to an area of 22 meters by 10 meters; the current position of the pedestrian agent will be placed between (-5, -12) and (5, -12); the desired destination of the agent will be placed between (-5, 10) and (5, 10).

The obstacle has a random chance of appearing in the environment. Each obstacle has a *size* ranged from 0.5 to 2 meters and a *danger level* ranged from 0 to 1.

The objective of our research is to let the pedestrian agent plan a path from its position to a pre-defined destination. In order to do this, the agent must observe the environment then provide a path using its current policy. In our model, the agent path is constructed from 10 outputs of the neural network, corresponding to 10 component path nodes' relative x positions. Appropriately, the component path nodes' relative y positions are {-10, -8, -6... 6, 8}.

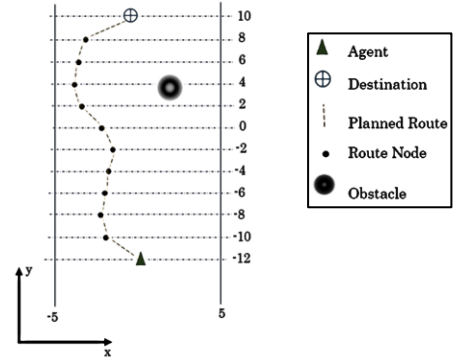

**Figure 2. Path-planning model.**

Specifically, for each training step, the pedestrian agent observes the following values:

- Relative x positions of agent's current position and its desired destination
- The presence of the obstacle. If the obstacle is present, the agent will observe the relative position, size and the danger level of the obstacle.

The taken action of the agent, which is the planned path in this case, will then be rewarded based on the rewarding behaviour discussed in Section 6. After that, the training step is terminated and the new training environment will be initialised.

## 6. REWARDING FORMULATION

As suggested in Section 4, we formulise our rewarding behaviour based on the idea of *human comfort.* There are many factors could affect human comfort level, but within the scope of research, we employed the following factors:

- Choosing the shortest path to the destination.
- Encouraging actions following the basic rules or common sense.
- Discouraging the changing of direction.
- Avoid getting through restricted regions.

Choosing the shortest path to the destination, as discussed in [5], is not always optimised for path planning but still has a very high priority in the process. For calculating the reward, we used the negative of the sum of all squared lengths of all walking paths. The bias $b$ used in the reward is for providing a positive reward to the agent when a satisfactory path is taken. The rewarding for taking the shortest path is formulated as follows

$$\mathcal{R}_1 = -\sum_{i=0}^{11} \|p_i\|^2 + b, \quad (1)$$

where $p_i$ is a vector representing the path from the previous node to the next node in the agent's planned path and $b$ is the bias.

For discouraging changing direction, we added a small penalty when a change in direction is greater than 30°. The reason for this is that in human navigation, a minor change in direction is still considered acceptable. The rewarding for changing direction Acknowledgements as follows

$$\mathcal{R}_2 = -\sum_{i=0}^{10} \theta(angle(p_i, p_{i+1}) - 30), \quad (2)$$

where $angle(p_i, p_j)$ is the value in degree of angle formed by two vectors $p_i$ and $p_j$; $\theta(x)$ is the Heaviside step function which is defined as

$$\theta(n) = \begin{cases} 0 \text{ if } n < 0 \\ 1 \text{ if } n \geq 0 \end{cases}.$$

In terms of following the basic rules or common sense, this could be varied depends on regional laws and cultures. In the scope of our research, we implemented the following principles:

(1) Favour going parallel to the sides. This will help the agent maintain the flow of the movement in the road.

(2) Following the left side of the road (or the right side, in case of right-side walking countries). Although pedestrians are not explicitly required by the laws to follow this convention in many countries, many people still follow the convention as a rule of thumb in the decision-making process in many situations.

(3) Avoid getting too close with the boundaries. As discussed in several studies, especially rule-based models, this is for avoiding accidental injuries when colliding with walls or surrounding objects. [3]

To implement these, we simulated the navigation along the path by sampling the planned path into N samples $s_i$, then calculate the appropriate rewards

$$\mathcal{R}_3 = -\sum_{i=0}^{N} \|x\_position(s_{i+1}) - x\_position(s_i)\|, \quad (3)$$

$$\mathcal{R}_4 = -\sum_{i=0}^{N} \theta(-x\_position(s_i)), \quad (4)$$

$$\mathcal{R}_5 = -\sum_{i=0}^{N} \theta(\|x\_position(s_i)\| - 4.5), \quad (5)$$

where $x\_position$ function is for getting the x coordinate of the position $s_i$.

Obstacle avoidance is probably the most essential criteria in path planning as it directly affects the pedestrian's safety. In real life, humans often try to keep a certain distance from the obstacle's centre, but once the distance is assured, the priority in the path planning process will shift to other interests. In the idea of reinforcement learning, when the path does not conflict with the obstacle area, a further distance from obstacle will not provide a higher reward. This idea was formulated in our rewarding for avoiding obstacle as follows:

$$\mathcal{R}_6 = \sum_{i=0}^{N} \begin{cases} \frac{\delta(s_i, obs)}{r_{obs}^2} * danger_{obs}^2 \text{ if } \delta(s_i, obs) < 0 \\ 0.01 * danger_{obs}^2 \text{ if } \delta(s_i, obs) > 0 \end{cases}, \quad (6)$$

with $\delta(s_i, obs) = d(s_i, obs)^2 - r_{obs}^2$,

where $d(s_i, obs)$ is the distance from the sampled position $s_i$ to obstacle's position; $r_{obs}$ and $danger_{obs}$ are the radius and the danger level of the obstacle area, respectively.

The cumulative reward is calculated by the sum of all component rewards mentioned above, each was multiplied by an appropriate coefficient:

$$\mathcal{R} = \sum_{i=1}^{6} \mathcal{R}_i * \kappa_i, \quad (7)$$

where $\kappa_i$ is the coefficient for rewarding of each reward.

These coefficients represent the proportion of importance of each reward, which can be different between agents. Variation of these coefficients could alternate the output results, and by that can be a representation of different human personalities. For example, a law-obedient pedestrian could use a high value for the coefficient of walking in the left side, while a cautious agent could use a high value for the coefficient of obstacle avoidance.

## 7. IMPLEMENTATION AND DISCUSSION

The realisation of our model was made available using Unity-ML [14], a framework which functions as a communicator between Python using TensorFlow and the 3D graphics engine Unity. For each training step, we initialise our environment then let our agent observes the current state. After that, these signals are sent to Python via the communicator for the training process. The output of the neural network using PPO algorithm will be then sent back to Unity and used for positioning the coordinates of each path node. The cumulative reward is calculated based on the output path and sent to Python for the training process.

We built the model for the environment entirely within Unity environment. The environment modelled is excessively noisy; therefore, a large size of batch is required. For faster training, we concurrently trained the model using 10 copies of the same environment. We have been able to successfully train the model with a batch size of 20480, buffer size of 204800 and the learning rate of 1.5e-3 in three million steps. As can be seen in **Figure 3** as follows, the reward has seemed to be converged at around -0.3.

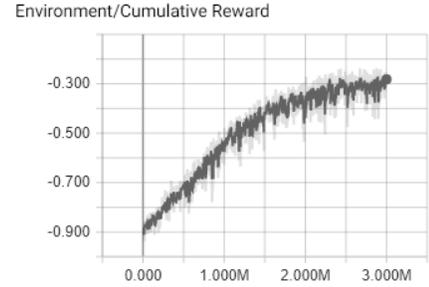

**Figure 3. Cumulative reward statistics.**

Using the trained model for pedestrian agent's path planning action, the behaviour can be observed from the results presented in **Figure 4**.

From observation, generally, it can be said that the agent's path resembles a similarity with a human person's decision of forming a walking path. In (a), the figure shows that the agent by our model planned a relatively short path that still conforms the walking convention such as walking on the left side of the road and changing direction naturally. On the contrary, the path formed by SFM leads the agent to go straight to the destination. In (b), there is an obstacle but outside of the agent's common planned path. The obstacle, in this case, has little to no effect on the result of its planned path; therefore, there are little changes to the path compared to the situation in (a). Similarly, no change was observed in the path formed by SFM as well. In (c), the obstacle now is in the agent's common planned path. In this case, the danger level of the obstacle perceived by the agent is very small, so our agent only tried to modify the path just enough to not conflict with the obstacle. As for the path by SFM, the agent still chooses to go straight to the destination and only try to avoid the obstacle when being close to it. When the danger level perceived was increased as in (d), our agent tried to stay away from the

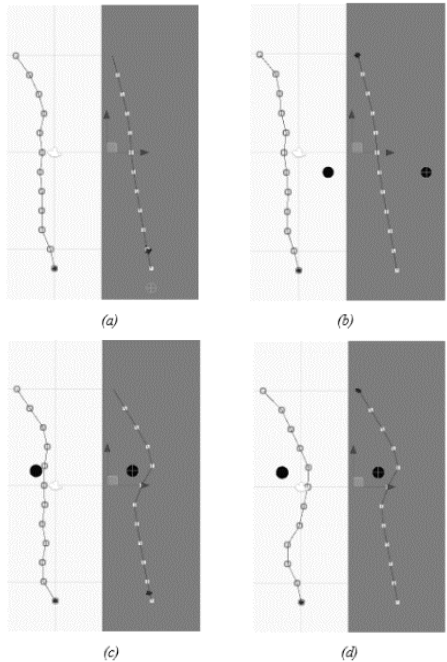

**Figure 4. Agent's planned path in different situations. The path using RL (our model) is on the left, the path using SFM is on the right.**

*(a) No obstacle; (b) Obstacle area outside agent's path; (c) Obstacle area within agent's path, danger level = 0.1; (d) Same obstacle area with danger level = 1.*

obstacle much further. As can be seen from the figure, there are parts of the planned path positioned slightly on the right side of the road. Also, the total length of the path is also not the shortest. This path has replicated the common behaviour from humans to ensure their highest safety while walking on the road. The danger level is not present in SFM, therefore there is no change to the path compared to the situation (c).

Compared to a force-based model, SFM in particular, our model has the advantage of replicating the natural behaviour of human navigation. Although a rule-based model, e.g. a finite state machine model, might be able to mimic the exact behaviour precisely, it is prone to have the limitation of the finite number of rulesets. Reinforcement learning, on the hands, has the advantage of creating diversity on human behaviour thanks to the shared concepts between neural network and reinforcement in machine learning and in real life.

However, our model is still in early-stage and require much further research in order to replicate a perfect pedestrian behaviour in multiple situations. The obstacle, for one, cannot represent a moving agent such as automobile or another pedestrian. The reason for that is when encountering with a moving obstacle, the agent needs different ways to plan a path. For instance, the agent might need to plan ahead by making predictions, as discussed in [14]; or adapt to the changes in the environment and make decisions synchronously. Another limitation of our research is the lack of changing in the agent's velocity. The variation in speed of pedestrians is also a major factor in replicating a natural walking behaviour. In the future, we will conduct further research to address these problems using the result presented in this paper as a groundwork.

## 8. CONCLUSION
We have proposed a novel reinforcement learning model for pedestrian agent path planning and collision avoidance. The implementation of our model has shown that the agent is able to plan a natural path to the destination while avoiding colliding with the obstacle in different situations. The planned path shares many similarities with a human pedestrian in several aspects such as following common walking conventions and human behaviours.